\documentclass[runningheads]{llncs}

\usepackage[utf8]{inputenc}
\usepackage[english]{babel}
\usepackage[T1]{fontenc}

\usepackage{amsmath}
\usepackage{amssymb}

\usepackage{graphicx}
\usepackage{svg}
\usepackage{float}

\usepackage{booktabs}
\usepackage{makecell}
\usepackage{multirow}
\usepackage{siunitx}

\usepackage{algorithm}
\usepackage{algorithmic}

\usepackage{hyperref}

\usepackage{xspace}
\usepackage{xcolor}
\usepackage{soul}
\usepackage{url}
\usepackage[inline]{enumitem}
\usepackage{multicol}
\usepackage[normalem]{ulem} 
\usepackage{pifont} 

\newtheorem{mytheorem}{Theorem}

\newcommand{\dcircle}[1]{\ding{\numexpr171 + #1}}

\newcommand{\toolname}{{LaFiCMIL}\xspace}

\usepackage{caption}

\begin{document}
\title{LaFiCMIL: Rethinking Large File Classification from the Perspective of Correlated Multiple Instance Learning}
%
%

\author{Tiezhu Sun\inst{1}\thanks{Corresponding Author, E-mail: \email{tiezhu.sun@uni.lu}} \and
Weiguo Pian\inst{1} \and
Nadia Daoudi\inst{2} \and Kevin Allix\inst{3} \and Tegawendé F. Bissyandé \inst{1} \and Jacques Klein \inst{1}}
\authorrunning{T. Sun et al.}
%
\institute{ University of Luxembourg, Kirchberg, Luxembourg \and Luxembourg Institute of Science and Technology, Esch-sur-Alzette, Luxembourg \and CentraleSupélec, Rennes, France}
\maketitle              
\begin{abstract}
Transfomer-based models have significantly advanced natural language processing, in particular the performance in text classification tasks. Nevertheless, these models face challenges in processing large files, primarily due to their input constraints, which are generally restricted to hundreds or thousands of tokens. 
Attempts to address this issue in existing models usually consist in extracting only a fraction of the essential information from lengthy inputs, while often incurring high computational costs due to their complex architectures.
In this work, we address the challenge of classifying large files from the perspective of correlated multiple instance learning. We introduce  LaFiCMIL, a method specifically designed for large file classification. It is optimized for efficient training on a single GPU, making it a versatile solution for binary, multi-class, and multi-label classification tasks.
We conducted extensive experiments using seven diverse and comprehensive benchmark datasets to assess  LaFiCMIL's effectiveness. By integrating BERT for feature extraction,  LaFiCMIL demonstrates exceptional performance, setting new benchmarks across all datasets. 
A notable achievement of our approach is its ability to scale BERT to handle nearly \num{20000} tokens while training on a single GPU with 32GB of memory.
This efficiency, coupled with its state-of-the-art performance, highlights  LaFiCMIL's potential as a groundbreaking approach in the field of large file classification.

\end{abstract}

\section{Introduction}
Text classification is a fundamental task in Natural Language Processing (NLP), entailing the assignment of suitable label(s) to specific input texts \cite{kowsari2019text,premasiri2023can}. This process is crucial across various domains, including sentiment analysis \cite{dang2020sentiment}, fake news detection \cite{kumar2020fake}, and offensive language identification \cite{ranasinghe2020multilingual}, among others. Recent years have seen the emergence of attention-based models like Transformer \cite{vaswani2017attention}, GPT \cite{radford2018improving,radford2019language}, and the BERT family \cite{devlin2018bert,feng2020codebert,sun2023dexbert}, which have established state-of-the-art benchmarks in text classification tasks. 
However, the challenge of processing very long sequences remains a significant obstacle, mainly due to their high computational requirements when facing extremely huge number of tokens. 

There are mainly two types of solutions in the literature to address long token sequences: \dcircle{1} extending the input length limit by employing a sliding window to attention, such as Longformer~\cite{beltagy2020longformer},
and \dcircle{2} dividing long documents into segments and recurrently processing the Transformer-based segment representations, such as RMT~\cite{bulatov2022recurrent,bulatov2023scaling}.
Nevertheless, Longformer inherently struggles with global context capture. The sliding window mechanism can lead to a fragmented understanding of the overall sequence, as it primarily focuses on local context within each window. This becomes particularly challenging when dependencies span beyond the scope of these localized windows, a common occurrence in complex or extremely long text sequences.
Similarly, the recurrent processing of RMT can also lead to information loss, especially when dependencies or context need to be carried over long sequences of text. Each recurrent step has the potential to dilute or overlook critical information from previous segments, leading to a gradual decay in context retention as the sequence progresses.

Recently, significant progress in large language models (LLMs) like GPT-4~\cite{achiam2023gpt}, ChatGLM~\cite{zeng2022glm}, and Llama 2~\cite{touvron2023llama} has showcased their impressive capabilities in various NLP tasks.
Despite these advances, directly applying large language models (LLMs) to text classification through prompt engineering has not achieved optimal performance, and lightweight Transformer models, such as RoBERTa~\cite{liu2019roberta,zhang2024pushing}, continue to excel in this essential task.
This is demonstrated by RGPT~\cite{zhang2024pushing}, an adaptive boosting framework designed to fine-tune LLMs for text classification.
Although RGPT sets new benchmarks, it has only been validated on short texts and demands substantial computational resources, requiring \textbf{eight} A100-SXM4-40GB GPUs, and a total of 320 GB of memory.
Consequently, developing a more resource-efficient and effective strategy for classifying long documents remains a critical challenge.

In this work, we innovatively tackle this challenge by adopting Multiple Instance Learning~\cite{shao2021transmil,zhang2022dtfd} (MIL), wherein we conceptualize a large file as a 'bag' and its constituent chunks as 'instances' within the MIL framework.
We introduce \textbf{\toolname}, a simple yet effective \textbf{La}rge \textbf{Fi}le \textbf{C}lassification approach based on correlated \textbf{M}ultiple \textbf{I}nstance \textbf{L}earning.
On the one hand, as proven in Theorem~\ref{theorem:approximation} (cf., Section~\ref{sec:preliminary}), a MIL score function for a bag classification task can be approximated by a series of sub-functions of the instances.
This inspires us to split a large file into smaller chunks and extract their features separately using BERT.
On the other hand, we aim to guide the model to learn high-level overall features from all instances, rather than deriving the final bag prediction from instance predictions based on a simplistic learned projection matrix.
In addition, in contrast to the basic version of MIL~\cite{ilse2018attention}, where instances within the same bag exhibit neither dependency nor ordering among one another, we claim that the small chunks from the same large file are correlated in some way (e.g., semantic dependencies in paragraphs). 
This implies that the presence or absence of a positive instance in a bag can be influenced by the other instances contained within the same bag.
As a result, relying on our computationally efficient LaFiAttention layer, our approach is capable of {\bf efficiently} extracting {\bf correlations} among {\bf all chunks} as additional information to boost classification performance. 

In our evaluation, \toolname consistently achieved new state-of-the-art performance across all seven benchmark datasets, especially when tested with long documents in the evaluation sets. A notable highlight is \toolname's performance on the full test set of the Paired Book Summary dataset, where it demonstrated a significant 4.41 percentage point improvement. This dataset is especially challenging as it contains the highest proportion of long documents, exceeding 75\%. Furthermore, \toolname also distinguished itself by having the fastest training process compared to other baseline models.

The contributions of our study are as follows: 
\begin{itemize}
    \item We introduce, \toolname, a novel approach for large file classification from the perspective of correlated multiple instance learning.
    \item The training of \toolname is super efficient, which requires only 1.86$\times$ training time than the original BERT, but is able to handle 39$\times$ longer sequence on a single GPU with only 32 GB of memory. 
    \item We perform a comprehensive evaluation, illustrating that \toolname achieves new state-of-the-art performance across all seven benchmark datasets.
    \item We share the datasets and source code to the community at: 
    \url{https://github.com/Trustworthy-Software/LaFiCMIL}
\end{itemize}

\section{Related Work}
\subsection{Large File Classification}
In recent years, significant efforts have been made to alleviate the input limit of Transformer-based models to handle large files.
One notable example is Longformer~\cite{beltagy2020longformer}, which extends the limit to \num{4096} tokens using a sparse attention mechanism~\cite{zaheer2020big}.
CogLTX~\cite{ding2020cogltx} chooses to identify key sentences through a trained judge model. 
Alternatively, ToBERT~\cite{pappagari2019hierarchical} and RMT~\cite{bulatov2022recurrent,bulatov2023scaling} segment long documents into fragments and then aggregate or recurrently process their BERT-based representations.
Recently, two simple BERT-based methods proposed in~\cite{park2022efficient} achieved state-of-the-art performance on several datasets for long document classification.
Specifically, BERT+Random selects random sentences up to 512 tokens to augment the first 512 tokens. 
BERT+TextRank augments the first 512 tokens with a second set of 512 tokens obtained via TextRank~\cite{mihalcea2004textrank}. 
They also provide a comprehensive evaluation to compare the relative efficacy of various baselines on diverse datasets, which revealed that no single approach consistently outperforms others across all six benchmark datasets, encompassing different classification tasks such as binary~\cite{kiesel2019semeval}, multi-class~\cite{lang1995newsweeder}, and multi-label classification~\cite{bamman2013new,chalkidis2019large}.

One potential reason for the limited performance of existing approaches is that they do not fully leverage the information available in large files, resulting in only partial essential information being captured. 
In this paper, we explore the possibility of utilizing the complete information from large files to improve the performance of various classification tasks.

\subsection{Multiple Instance Learning}
Multiple Instance Learning (MIL) has attracted increasing research interests and applications in recent years. 
The application scenarios of MIL span across various domains~\cite{ji2020diversified,song2019using,hebbar2021deep}, but the most prominent one is Medical Imaging and Diagnosis.
Particularly, there has been a growing trend towards developing MIL algorithms for medical whole slide image analysis~\cite{kanavati2020weakly,xu2019camel}.
MIL approaches are broadly divided into two categories. The first group makes bag predictions based on individual instance predictions, typically using average or maximum pooling methods~\cite{feng2017deep,lerousseau2020weakly,lu2021data,sharma2021cluster}. 
The second group aggregates instance features to form a high-level bag representation, which is then used for bag-level predictions~\cite{li2021dual,shao2021transmil,zhang2022dtfd}. 
Although instance-level pooling is straightforward, aggregating features for bag representation has proven more effective~\cite{wang2018revisiting,shao2021transmil}. 

A fundamental assumption behind Multiple Instance Learning (MIL) is that instances within a bag are independent, but this might not always be true in real world.
To tackle this, some research turns to Correlated Multiple Instance Learning (c-MIL), which assumes instances in a bag are correlated~\cite{zhou2009multi,zhang2021non,shao2021transmil}.
This approach recognizes that instances in a bag can influence each other. However, using c-MIL for large file classification remains under-explored.
\section{Technical Preliminaries}
\label{sec:preliminary}
In this section, we describe several essential technical preliminaries that inspire and underpin the design of \toolname.
We first present a pair of theorems that are the fundamental principles of c-MIL.

\begin{mytheorem}
\label{theorem:approximation}
Suppose $S : \chi \rightarrow \mathbb{R}$ is a continuous set function w.r.t Hausdorff distance~\cite{rote1991computing} $d_{H}(., .)$. $\forall \varepsilon > 0$, for any invertible map $P : \chi \rightarrow \mathbb{R}^{n}$, $\exists$ function $\sigma$ and $g$, such that for any set $X \in \chi$:

\begin{equation}
\small
    |S(X) - g(P_{X \in \chi}\{\sigma(x): x \in X\})| < \varepsilon
\end{equation}
\end{mytheorem}

The proof of Theorem~\ref{theorem:approximation} can be found in~\cite{shao2021transmil}.
This theorem shows that a Hausdorff continuous {\bf set function} $S(X)$ can be arbitrarily approximated by a function in the form $g(P_{X \in \chi}\{\sigma(x) : x \in X\})$.
This insight can be applied to MIL, as the mathematical definition of {\bf sets} in the theorem is equivalent to that of \textbf{bags} in MIL framework. 
Consequently, the theorem provides a foundation for approximating bag-level predictions in MIL using instance-level features.

\begin{mytheorem}
\label{theorem:entropy}
The instances in the bag are represented by random variables $\theta_{1}, \theta_{2}, ..., \theta_{n}$, the information entropy of the bag under the correlation assumption can be expressed as $H(\theta_{1}, \theta_{2}, ..., \theta_{n})$, and the information entropy of the bag under the i.i.d. (independent and identical distribution) assumption can be expressed as $\sum_{t=1}^{n} H(\theta_{t})$, then we have:
\begin{equation}
\small
\begin{split}
H(\theta_{1}, \theta_{2}, ..., \theta_{n}) &= \sum_{t=2}^{n} H(\theta_{t} | \theta_{1}, \theta_{2}, ..., \theta_{t-1}) + H(\theta_{1}) \\ & \leq \sum_{t=1}^{n} H(\theta_{t})
\end{split}
\end{equation}
\end{mytheorem}
The proof of Theorem~\ref{theorem:entropy} can be found in~\cite{shao2021transmil}.
This theorem demonstrates that the information entropy of a bag under the correlation assumption is smaller than the information entropy of a bag under the i.i.d. assumption. 
The lower information entropy in c-MIL suggests reduced uncertainty and the potential to provide more information for bag classification tasks.
In Section~\ref{sec:corremil}, we introduce c-MIL, and in Section~\ref{sec:laficMIL}, we derive the efficient \toolname based on c-MIL. 

Next, we present the necessary preliminaries for our efficient attention layer inspired by the Nyströmformer~\cite{xiong2021nystromformer}, referred as LaFiAttention, which performs as a sub-function within our \toolname.
In the original Transformer~\cite{vaswani2017attention}, an input sequence of $n$ tokens of dimensions $d$, $X \in {\bf R}^{n \times d}$, is projected using three matrices $W_{Q} \in {\bf R}^{n \times d_{q}}$, $W_{K} \in {\bf R}^{n \times d_{k}}$, and $W_{V} \in {\bf R}^{n \times d_{v}}$, referred as query, key, and value respectively with $d_{k} = d_{q}$. The outputs $Q, K, V$ are calculated as

\begin{equation}
\small
    Q = XW_{Q},~K = XW_{K},~V = XW_{V}
\end{equation}
Therefore, the self-attention can be written as:
\begin{equation}
\small
    D(Q, K, V) = SV = softmax(\frac{QK^T}{\sqrt{d_{q}}})V
\end{equation}
Then, the softmax matrix $S$ used in self-attention can be written as

\begin{equation}
\small
    S = softmax(\frac{QK^T}{\sqrt{d_{q}}}) = \begin{bmatrix}
        A_{S} & B_{S} \\ 
        F_{S} & C_{S}
    \end{bmatrix}
\end{equation}
where $A_{S} \in {\bf R}^{m \times m}$, $B_{S} \in {\bf R}^{m \times (n - m)}$, $F_{S} \in {\bf R}^{(n - m) \times m}$, $C_{S} \in {\bf R}^{(n - m) \times (n - m)}$, and $m < n$.

In order to {\bf reduce} the memory and time {\bf complexity} from $O(n^{2})$ to $O(n)$, LaFiAttention approximates $S$ by 
\begin{equation}
\small
    \hat{S} = softmax(\frac{Q\widetilde{K}^T}{\sqrt{d_{q}}})A^{+}_{S}softmax(\frac{\widetilde{Q}K^T}{\sqrt{d_{q}}}),
\end{equation}
where $\widetilde{Q} = [\widetilde{q_{1}};...;\widetilde{q_{m}}] \in {\bf R}^{m \times d_{q}}$ and $\widetilde{K} = [\widetilde{k_{1}};...;\widetilde{k_{m}}] \in {\bf R}^{m \times d_{q}}$ are the selected landmarks for inputs $Q = [q_{1};...;q_{n}]$ and $K = [k_{1};...;k_{n}]$, $A^{+}_{S}$ is the Moore-Penrose inverse~\cite{ben2003generalized} of $A_{S}$. 

\begin{lemma}
\label{lemma:moore-penrose}
For $A_{S}\in {\bf R}^{m\times m}$, the sequence $\{Z_{j}\}^{j=\infty}_{j=0}$ generated by~\cite{razavi2014new},
\begin{equation}
\small
    Z_{j+1} = \frac{1}{4}Z_{j}(13I - A_{S}Z_{j}(15I - A_{S}Z_{j}(7I - A_{S}Z_{j})))
\end{equation}
converges to Moore-Penrose inverse $A^{+}_{S}$ in the third-order with initial approximation $Z_{0}$ satisfying $\left \| A_{S}A^{+}_{S} - A_{S}Z_{0} \right \| < 1$.
\end{lemma}

LaFiAttention approximates $A^{+}_{S}$ by $Z^{*}$ with Lemma~\ref{lemma:moore-penrose}.
Following the empirical choice from~\cite{xiong2021nystromformer}, we run 6 iterations to achieve a good approximation of the pseudoinverse.
Then, the softmax matrix $S$ is approximated by

\begin{equation}
\small
\label{eq:nystrom}
    \hat{S} = softmax(\frac{Q\widetilde{K}^T}{\sqrt{d_{q}}})Z^{*}softmax(\frac{\widetilde{Q}K^T}{\sqrt{d_{q}}}).
\end{equation}

\section{Approach}
\label{sec:approach}

In this section, we first introduce customized c-MIL for large file classification and then provide technical details about our \toolname approach.

\begin{figure*}[htp]
\centering
\includegraphics[width=.98\linewidth]{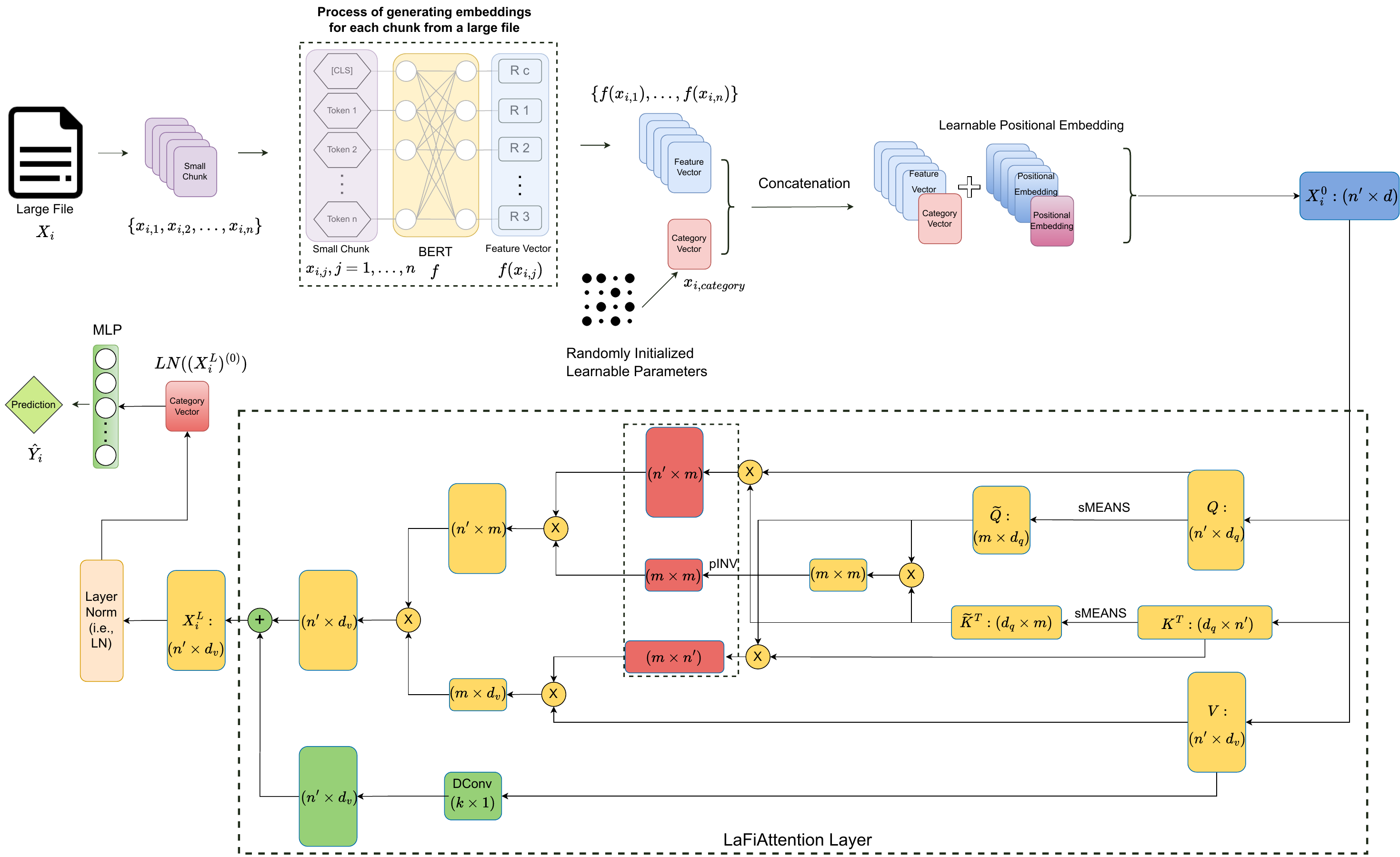}
\caption{
\toolname. Initially, document chunks are transformed into embedding vectors using BERT. A learnable category vector is then concatenated to these embeddings to form an augmented bag $X_i^0$ with $n' = n + 1$ instances. The LaFiAttention layer captures the inter-instance correlations within $X_i^0$. Operations within this layer, such as matrix multiplication ($\times$) and addition ($+$), are specified alongside the variable names and matrix dimensions. Key processes include sMEANS for landmark selections similar to~\cite{shen2018baseline}, pINV for pseudoinverse approximation, and DConv for depth-wise convolution. Classification is completed by passing the learned category vector through a fully connected layer.}
\label{fig:laficmil}
\end{figure*}

\subsection{Correlated Multiple Instance Learning}
\label{sec:corremil}
Unlike traditional supervised classification, which predicts labels for individual instances, Multiple Instance Learning (MIL) predicts bag-level labels for bags of instances. 
Typically, individual instance labels within each bag exist but inaccessible, and the number of instances in different bags may vary.

In the basic MIL concept~\cite{ilse2018attention},  instances in a bag are independent and unordered. 
However, correlations among chunks of a large file exist due to the presence of semantic dependencies between paragraphs.
According to Theorem~\ref{theorem:entropy}, these correlations can be exploited to reduce uncertainty in prediction. 
In other words, this relationship can be leveraged as additional information to boost the performance of long document classification tasks.
The Correlated Multiple Instance Learning (c-MIL) is defined as below.

Here, we consider a binary classification task of c-MIL as an example.
Given a bag (i.e., a large file) $X_{i}$ composed of instances (i.e., chunks) $\{x_{i,1}, x_{i,2}, ..., x_{i,n}\}$, for $i=1, ..., N$, that exhibit dependency or ordering among each other.
The bag-level label is $Y_{i}$, yet the instance-level labels $\{y_{i,1}, y_{i,2}, ..., y_{i,n}\}$ are not accessible.
Then, a binary classification of c-MIL can be defined as:
\begin{equation}
\small
Y_{i} = \left\{\begin{array}{l}
0, \indent if \sum y_{i,j}=0 \indent y_{i,j}\in \{0, 1\}, j=1,...,n\\ 
1, \indent otherwise
\end{array}\right.
\end{equation}

\begin{equation}
\small
\label{eq:score}
    \hat Y_{i} = S(X_{i}),
\end{equation}
where $S$ is a scoring function, and $\hat Y$ is the predicted score. 
$N$ is the total number of bags, and $n$ is the number of instances in the $i$th bag. 
The number $n$ generally varies for different bags.

\subsection{\toolname}
\label{sec:laficMIL}
According to Theorem~\ref{theorem:approximation}, we leverage Multi-layer Perceptron~\cite{rumelhart1986learning}, BERT~\cite{devlin2018bert}, LaFiAttention Layer and Layer Normalization~\cite{ba2016layer} as {\bf sub-functions to approximate} the c-MIL score function $S$ defined in Equation~\ref{eq:score}. 

Given a set of bags $\{X_{1}, ..., X_{N}\}$, where each bag $X_{i}$ contains multiple instances $\{x_{i,1}, ..., x_{i,n}\}$, a bag label $Y_{i}$, and a randomly initialized category vector $x_{i, category}$, the goal is to learn the maps: $\mathbb{X} \rightarrow \mathbb{T} \rightarrow \gamma$, where $\mathbb{X}$ is the bag space, $\mathbb{T}$ is the transformer space and $\gamma$ is the label space. The map of $\mathbb{X} \rightarrow \mathbb{T}$ can be defined as:
\begin{equation}
\small
\begin{split}
\label{eq:vecs}
    X_{i}^{0} = & [x_{i, category}; f(x_{i,1}); ...; f(x_{i,n})] \\ & + E_{pos}, \indent X_{i}^{0}, E_{pos} \in \mathbb{R}^{(n+1) \times d}
\end{split}
\end{equation}
\begin{equation}
\small
\begin{split}
    Q^{l} = X_{i}^{l-1}W_{Q}, & \indent K^{l} = X_{i}^{l-1}W_{K}, \indent V^{l} = X_{i}^{l-1}W_{V}, \\ & \indent l=1, ..., L
\end{split}
\end{equation}
where function $f$ is approximated by a BERT model, $E_{pos}$ is the Positional Embedding, and $L$ is the number of Multi-head Self-Attention ($MSA$) block. 
\begin{equation}
\small
\begin{split}
    head & = LaFiSA(Q^{l}, K^{l}, V^{l}) \\ & = softmax(\frac{Q^{l}(\widetilde{K}^{l})^T}{\sqrt{d_{q}}})Z^{*l}softmax(\frac{\widetilde{Q}^{l}(K^{l})^T}{\sqrt{d_{q}}})V^{l}, \\ 
\end{split}
\end{equation}
\begin{equation}
\small
    MSA(Q^{l}, K^{l}, V^{l}) = Concat(head_{1}, ..., head_{h})W_{O}, 
\end{equation}
\begin{equation}
\small
    X_{i}^{l} = MSA(LN(X_{i}^{l-1})) + X_{i}^{l-1}, \indent l=1, ..., L
\end{equation}
where $W_{O} \in \mathbb{R}^{hd_{v} \times d}$, $head \in \mathbb{R}^{(n+1) \times d_{v}}$, $LaFiSA$ denotes the approximated Self-attention layer by Nyström method~\cite{baker1977numerical} according to Equation~\ref{eq:nystrom}, $h$ is the number of head in each $MSA$ block, and Layer Normalization(LN) is applied before each $MSA$ block. 

The map of $\mathbb{T} \rightarrow \gamma$ can be simply defined as:
\begin{equation}
\small
   \hat Y_{i} = MLP(LN((X_{i}^{L})^{(0)})),
\end{equation}
where $(X_{i}^{L})^{(0)}$ represents the learned category vector, and $MLP$ means Multi-layer Perceptron (i.e., fully connected layer).

From the above formulation, we can find that the most important part is to efficiently learn the map from bag space $\mathbb{X}$ to Transformer space $\mathbb{T}$.
As illustrated in Figure~\ref{fig:laficmil}, this map is approximated by a series of sub-functions which are approximated by various neural layers.
The overall process is summarized as follows: given a large file, we use a BERT model to generate the representations of the divided chunks (i.e., instances in the concept of c-MIL). 
Then, we initialize a {\bf learnable} category vector that follows a normal distribution and has the same shape as each instance.
By considering the category vector as an additional instance, we learn the correlation between each instance using LaFiAttention layer.
With the help of the attention mechanism, the category vector exchanges information with each chunk and extracts necessary features for large file classification.
Finally, the category vector is fed into a fully connected layer to finalize the classification task.

\section{Experimental Setup}


\begin{table}
  \centering
  \caption{Statistics on the datasets. \# BERT Tokens indicates the average token number obtained via the BERT tokenizer. \% Long Docs means the proportion of documents exceeding 512 BERT tokens.}
  \tiny 
  \setlength{\tabcolsep}{4pt} 
  \begin{tabular}{@{}lcccccSSS@{}} 
    \toprule
    Dataset & Type & {\# Total} & {\# Train} & {\# Val} & {\# Test} & {\# Labels} & {\# BERT Tokens} & {\% Long Docs} \\
    \midrule
    Hyperpartisan & binary & 645 & 516 & 64 & 65 & 2 & {744.18$\pm$677.87} & 53.49 \\
    20NewsGroups & multi-class & 18846 & 10182 & 1132 & 7532 & 20 & {368.83$\pm$783.84} & 14.71 \\
    Book Summary & multi-label & 12788 & 10230 & 1279 & 1279 & 227 & {574.31$\pm$659.56} & 38.46 \\
    -Paired & multi-label & 6393 & 5115 & 639 & 639 & 227 & {1148.62$\pm$933.97} & 75.54 \\
    EURLEX-57K & multi-label & 57000 & 45000 & 6000 & 6000 & 4271 & {707.99$\pm$538.69} & 51.3 \\
    -Inverted & multi-label & 57000 & 45000 & 6000 & 6000 & 4271 & {707.99$\pm$538.69} & 51.3 \\
    Devign & binary & 27318 & 21854 & 2732 & 2732 & 2 & {615.46$\pm$41917.54} & 39.76 \\
    \bottomrule
  \end{tabular}
  \label{tab:stat}
\end{table}


\begin{table}[htbp]
  \centering
  \scriptsize
  \caption{Performance metrics on only long documents in test set. The highest score is bolded and underlined, while the second highest score is only bolded. The subsequent tables of this task are organized in a consistent manner.} 
  \begin{tabular}{lcccccc}
    \toprule
    Model & Hyperpartisan & 20News & EURLEX & -Inverted & Book & -Paired \\
    \midrule
    BERT  & 88.00 & 86.09 & 66.76 & 62.88 & 60.56 & 52.23 \\
    -TextRank & 85.63 & 85.55 & 66.56 & 64.22 & 61.76 & 56.24 \\
    -Random & 83.50 & \textbf{86.18} & \textbf{67.03} & \textbf{64.31} & \textbf{62.34} & 56.77 \\
    Longformer & \textbf{93.17} & 85.50 & 44.66 & 47.00 & 59.66 & \textbf{58.85} \\
    ToBERT & 86.50 & -     & 61.85 & 59.50 & 61.38 & 58.17 \\
    CogLTX & 91.91 & 86.07 & 61.95 & 63.00 & 60.71 & 55.74 \\
    RMT & 90.04 & 83.62 & 64.16 & 63.21 & 60.62 & 58.27 \\
    \midrule
    \toolname & \underline{\textbf{95.00}} & \underline{\textbf{87.49}} & \underline{\textbf{67.28}} & \underline{\textbf{65.04}} & \underline{\textbf{65.41}} & \underline{\textbf{63.03}} \\
    \bottomrule
  \end{tabular}
  \label{tab:text_long_res}%
\end{table}%

\textbf{Datasets.}
To ensure a fair comparison with baselines, we adopt the same benchmark datasets utilized in the state-of-the-arts for long document classification~\cite{park2022efficient}. 
We first evaluate \toolname on these six benchmark datasets: 
\dcircle{1} Hyperpartisan~\cite{kiesel2019semeval}, a compact dataset encompassing 645 documents, designed for \textit{binary classification}. 
\dcircle{2} 20NewsGroups~\cite{lang1995newsweeder}, comprising 20 balanced categories and \num{11846} documents.
\dcircle{3} CMU Book Summary~\cite{bamman2013new}, tailored for \textit{multi-label classification}, contains \num{12788} documents and 227 genre labels. 
\dcircle{4} Paired Book Summary~\cite{park2022efficient}, formulated by combining pairs of documents from the CMU Book Summary dataset, features longer documents. 
\dcircle{5} EURLEX-57K~\cite{chalkidis2019large}, a substantial \textit{multi-label classification} dataset consisting of \num{57000} EU legal documents and \num{4271} available labels. 
\dcircle{6} Inverted EURLEX-57K~\cite{park2022efficient}, a modified version of EURLEX-57K dataset in which the order of sections is inverted, ensuring that core information appears towards the end of the document. 
To evaluate our method's effectiveness with longer documents, we incorporated the Devign dataset~\cite{zhou2019devign} for code defect detection. It features documents that are substantially longer than those in the other six datasets, with lengths approaching or exceeding 20,000 tokens.
Table~\ref{tab:stat} provides details of the datasets, covering metrics such as the average token count and its standard deviation, the maximum and minimum token counts, and the percentage of large documents, etc.

\textbf{Implementation Details.}
We split a long text document into chunks (i.e., c-MIL instances), and follow the standard BERT input length (i.e., 512 tokens) for each chunk.
To ensure a fair comparison, in line with the baselines, we employ BERT~\cite{devlin2018bert} for the first six datasets, and CodeBERT~\cite{feng2020codebert} and VulBERTa~\cite{hanif2022vulberta} for Devign, as the feature extractor.
Since we treat all chunks in each long document as a mini-batch, the actual batch size varies depending on the number of chunks in the long document.
We construct LaFiAttention layer with eight attention heads.
Under these configurations, a single Tesla V-100 GPU with 32GB of memory on an NVIDIA DGX Station can fully process 100\% of the large documents in the first six benchmark datasets and 99.92\% of those from Devign.
As a result, the average inference time (0.026s) of each mini-batch is almost the same as BERT (0.022s).
During training, the Adam optimizer~\cite{kingma2014adam} is leveraged.
The loss function varies depending on specific task.
Following the baseline~\cite{park2022efficient}, we use sigmoid and binary cross entropy for binary and multi-label classification, and softmax and cross entropy loss for multi-class classification.
For the Hyperpartisan, Book Summary, EURLEX-57K, and Devign datasets, a learning rate of \texttt{5e-6} is used, while \texttt{5e-7} is applied for the 20NewsGroups dataset.
We fine-tune the model for 10, 20, 40, 60, and 100 epochs on Devign, Hyperpartisan, 20NewsGroups, EURLEX-57K, and Book Summary, respectively.


\begin{table}[htbp]
  \centering
  \scriptsize
  \caption{Performance metrics on complete test set. The highest score in each column is bolded and underlined, while the second highest score is only bolded.}
  \begin{tabular}{lcccccc}
    \toprule
    Model & Hyperpartisan & 20News & EURLEX & -Inverted & Book & -Paired \\
    \midrule
    BERT  & 92.00 & 84.79 & 73.09 & 70.53 & 58.18 & 52.24 \\
    -TextRank & 91.15 & 84.99 & 72.87 & 71.30  & 58.94 & 55.99 \\
    -Random & 89.23 & 84.65 & \textbf{73.22} & \textbf{71.47} & \textbf{59.36} & 56.58 \\
    Longformer & \textbf{95.69} & 83.39 & 54.53 & 56.47 & 56.53 & \textbf{57.76} \\
    ToBERT & 89.54 & \underline{\textbf{85.52}} & 67.57 & 67.31 & 58.16 & 57.08 \\
    CogLTX & 94.77 & 84.62 & 70.13 & 70.80  & 58.27 & 55.91 \\
    RMT & 94.34 & 82.87 & 71.46 & 70.99  & 57.30 & 56.95 \\
    \midrule
    \toolname & \underline{\textbf{96.92}} & \textbf{85.07} & \underline{\textbf{73.72}} & \underline{\textbf{72.03}} & \underline{\textbf{61.34}} & \underline{\textbf{62.17}} \\
    \bottomrule
  \end{tabular}
  \label{tab:text_full_res}%
\end{table}%

\textbf{Evaluation Setup.}
We evaluate the performance of \toolname using the same metrics as those employed in the baselines~\cite{park2022efficient,hanif2022vulberta}.
We report the accuracy (\%) for binary and multi-class classification. 
We use micro-F1 (\%) for multi-label classification, 
and detection accuracy (\%) for code defect detection.
The results are the average of five independent runs, each with a different random seed.
As we explained in the introduction, our work aims to address this challenge by focusing on using a more practical resource: a single GPU with 32 GB of memory, which is only a tenth of what RGPT demands (i.e., 320 GB). 
Therefore, given the distinct resource requirements and application contexts, comparing our approach with LLMs falls outside the scope of this study.

\section{Experimental Results}
In this section, we analyze the performance of \toolname in long document classification.
We first discuss the overall performance, followed by an computational efficiency analysis and an ablation study on the core concepts of \toolname.

\subsection{Overall Performance}
Our experimental results reveal a phenomenon similar to ~\cite{park2022efficient} in that no existing approach consistently outperforms the others across all benchmark datasets.
However, as shown in Table~\ref{tab:text_long_res}, our \toolname establishes new state-of-the-art performance on all six NLP benchmark datasets when considering only long documents in the test set. 
Here, we define a long document as one containing over 512 BERT tokens. 
As shown in Table~\ref{tab:text_full_res}, we also achieve new state-of-the-art performance on five out of six NLP datasets when considering the full data (i.e., a mix of long and short documents) in the test set.
Particularly, we significantly improve the state-of-the-art score from 57.76\% to 62.17\% on the Paired Book Summary dataset, which contains the highest proportion of long documents (i.e., more than 75\%).
In contrast, we fail to achieve the best performance on 20NewsGroups, as the proportion of long documents in this dataset is very small (only 14.71\%); thus, our improvement on {\bf long} documents (as shown in Table~\ref{tab:text_long_res}) cannot dominate the overall performance on the entire dataset. 
This phenomenon is consistent with our motivation that the more long documents present in the dataset, the more correlations \toolname can extract to boost classification performance.


\begin{table}[ht]
\centering
\scriptsize
\caption{Accuracy (\%) comparison of different models on Devign dataset for code defect detection. The highest accuracy score is bolded and underlined, and the base model results are only bolded.}
\begin{tabular}{ccccccc}
\toprule
RoBERTa & CodeBERT & Code2vec~\cite{alon2019code2vec} & PLBART~\cite{ahmad2021unified} & VulBERTa & \makecell{CodeBERT+\\LaFiCMIL} & \makecell{VulBERTa+\\LaFiCMIL} \\
\midrule
61.05 & \textbf{62.08} & 62.48 & 63.18 & \textbf{64.27} & 63.43 & \underline{\textbf{64.53}} \\
\bottomrule
\end{tabular}
\label{tab:defect}
\end{table}

\begin{table}[htbp]
\centering
\scriptsize
\caption{Runtime and memory requirements of each model, {\bf relative to BERT}, based on the Hyperpartisan dataset. Training and inference time were measured and compared in seconds per epoch. GPU memory requirement is in GB.} 
\begin{tabular}{lccc}
\toprule
Model       & Train Time & Inference Time & GPU Memory \\
\midrule
BERT        & 1.00       & 1.00           & <16        \\
-TextRank   & 1.96       & 1.96           & 16         \\
-Random     & 1.98       & 2.00           & 16         \\
Longformer  & 12.05      & 11.92          & 32         \\
ToBERT      & 1.19       & 1.70           & 32         \\
CogLTX      & 104.52     & 12.53          & <16        \\
RMT         & 2.95       & 2.87           & 32        \\
\midrule
LaFiCMIL    & 1.86       & 1.18           & <32        \\
\bottomrule
\end{tabular}
\label{tab:efficiency}
\end{table}

Given that 100\% of long documents from the six NLP datasets can be fully processed, we conduct an additional evaluation of \toolname's ability to process {\bf extremely long} sequences, based on the code defect detection dataset Devign. Our findings reveal that LaFiCMIL is capable of handling inputs of up to nearly \num{20000} tokens when utilizing CodeBERT~\cite{feng2020codebert} and VulBERTa~\cite{hanif2022vulberta} as feature extractors on a {\bf single} GPU setup. This capability allows for 99.92\% of the code files in the Devign dataset to be processed in their entirety.
Concurrently, as shown in Table~\ref{tab:defect}, LaFiCMIL enhances the performance of both CodeBERT and VulBERTa, establishing a new state-of-the-art in accuracy over the evaluated baselines. 
We find that code defect detection is a challenging task on which most existing state-of-the-art models struggle to achieve even a single percentage point improvement over previous models.
Nonetheless, our \toolname helps CodeBERT gain 1.35 percentage points, representing a significant improvement.

\subsection{Computational Efficiency Analysis}
In this section, we provide a comprehensive analysis of computational efficiency outlined in Table~\ref{tab:efficiency}.
All models were evaluated on a single GPU with 32GB of memory using the Hyperpartisan dataset. 
LaFiCMIL performs distinctly in this context, demonstrating a runtime nearly on par with BERT.
The balance between high computational efficiency and advanced classification capability illustrates LaFiCMIL's exceptional capability to efficiently process long documents without significant computational overhead.

\subsection{Ablation Study}

\begin{table}[htbp]
  \centering
  \scriptsize
  \caption{Concept ablation study on long documents in test set. "wo" means "\toolname~{\bf w}ith{\bf o}ut".}
  \begin{tabular}{lcccccc}
    \toprule
    Model & Hyperpartisan & 20News & EURLEX & -Inverted & Book & -Paired \\
    \midrule
    wo BERT  & 85.00 & 53.92 & 60.54 & 54.14 & 50.11 & 46.61 \\
    wo LaFiAttn  & 87.50 & 84.97 & \textbf{66.82} & \textbf{64.89} & \textbf{62.50} & \textbf{60.13} \\
    wo c-MIL  & \textbf{88.00} & \textbf{86.09} & 66.76 & 62.88 & 60.56 & 52.23 \\
    \midrule
    \toolname & \underline{\textbf{95.00}} & \underline{\textbf{87.22}} & \underline{\textbf{67.28}} & \underline{\textbf{65.04}} & \underline{\textbf{65.41}} & \underline{\textbf{63.03}} \\
    \bottomrule
  \end{tabular}
  \label{tab:concept_ablation}%
\end{table}%

To gain a comprehensive understanding of the efficacy of each core concept in our approach (namely, BERT, LaFiAttention, and c-MIL), we conduct an ablation study. 
This study aims to evaluate the classification performance of \toolname when each concept is systematically removed, allowing us to evaluate their individual contributions.

Without a feature extractor, any approach would be ineffective. Thus, when BERT is removed, the LaFiAttention layer must assume the role of feature extractor instead of c-MIL. 
This would result in the disappearance of the c-MIL mechanism, and the approach can now only take the first chunk as input, transforming it into a basic attention-based classifier. 
As might be expected, the absence of the BERT concept leads to the worst performance across all datasets among the three variants, as shown in the first row of Table~\ref{tab:concept_ablation}.
If excluding the LaFiAttention concept, c-MIL devolves into a standard MIL, for which we employ the widely accepted Attention-MIL~\cite{ilse2018attention}. 
The results of this setting are presented in the second row of Table~\ref{tab:concept_ablation}. Given that this variant can still process all chunks of a lengthy document, it performs best among all three variants on the four datasets with the largest number and longest length of documents.
When the c-MIL concept is removed, the LaFiAttention layer will also be absent as it executes c-MIL which is no longer needed, leaving only BERT.
Due to its restriction to process only the first chunk as input, this variant fails to achieve the best results on the four datasets that are dominated by long documents.
Finally, upon comparing the three variants with the full \toolname, shown in the fourth row of Table~\ref{tab:concept_ablation}, it becomes evident that the exclusion of any concept significantly weakens performances across all datasets.

\section{Conclusion}
We propose \toolname, a large file classification approach based on correlated multiple instance learning.
Our method treats large document chunks as c-MIL instances, enabling feature extraction for classification from correlated chunks without substantial information loss.
Experimental results demonstrate that our approach significantly outperforms the state-of-the-art baselines across multiple benchmark datasets in terms of both efficiency and accuracy.
Our work provides a new perspective for addressing the large document classification problem.


%
%
%
\bibliographystyle{splncs04}
%
\bibliography{references}

\end{document}